%% file: emnlp2018.tex
\documentclass[11pt,a4paper]{article}
\usepackage[hyperref]{emnlp2018}
\usepackage{times}
\usepackage{latexsym}
\usepackage{amsmath}
\usepackage{url,comment}
\usepackage{array,multirow,booktabs}

\aclfinalcopy 

\newcommand{\jacob}[1]{\textcolor{purple}{[#1 -J]}}

\title{Pre-trained Sentence Embeddings for Implicit Discourse Relation Classification}

\author{Murali Raghu Babu Balusu \\
  School of Interactive Computing \\
  Georgia Institute of Technology \\
  Atlanta, GA, USA \\
  {\tt b.murali@gatech.edu} \\
  \And
  Yangfeng Ji \\
  Paul G. Allen School of \\
  Computer Science and \\
  Engineering\\
  University of Washington \\
  Seattle, WA, USA \\
  {\tt yangfeng@} \\
  {\tt cs.washington.edu} \\
  \And
  Jacob Eisenstein \\
  School of Interactive Computing \\
  Georgia Institute of Technology \\
  Atlanta, GA, USA \\
  {\tt jacobe@gatech.edu} \\}
  
\date{}

\begin{document}
\maketitle
\begin{abstract}
\input{abstract}

\end{abstract}

\input{intro}

\input{data}

\input{methods}

\input{results}

\input{relatedwork}

\input{conclusion}

\bibliography{emnlp2018}
\bibliographystyle{acl_natbib_nourl}


\end{document}

%% file: abstract.tex
Implicit discourse relations bind smaller linguistic units into coherent texts. Automatic sense prediction for implicit relations is hard, because it requires understanding the semantics of the linked arguments. Furthermore, annotated datasets contain relatively few labeled examples, due to the scale of the phenomenon: on average each discourse relation encompasses several dozen words.
In this paper, we explore the utility of pre-trained sentence embeddings as base representations in a neural network for implicit discourse relation sense classification.
We present a series of experiments using both supervised end-to-end trained models and pre-trained sentence encoding techniques --- SkipThought, Sent2vec and Infersent.
The pre-trained embeddings are competitive with the end-to-end model, and the approaches are complementary, with combined models yielding significant performance improvements on two of the three evaluations.



%% file: intro.tex
\section{Introduction}

Discourse relations describe the high-level organization of text. Identifying these discourse relations has been shown to be helpful to many downstream applications, such as sentiment analysis~\cite{Somasundaran:2009:SUM:1699510.1699533}, summarization~\cite{DBLP:conf/emnlp/YoshidaSHN14,Louis:2010:DIC:1944506.1944533}, question answering \cite{Jansen:14}, and coherence evaluation~\cite{Lin:2011:AET:2002472.2002598}. But
discourse relations that are not explicitly marked by connectives are difficult to classify, with state-of-the-art performance slightly above 40\%~\cite{ji2015one,Lin:2009:RID:1699510.1699555} on the Penn Discourse Treebank dataset~\cite{rashmi2008penn}.

One hypothesis is that poor performance is due to two factors: limited availability of training data and the semantic nature of the task. If so, then recent work on \emph{sentence embeddings} might help. These methods propose to embed sentences into a vector representation, and have been shown to be helpful for an array of tasks that appear to share some of the semantic characteristics of discourse relation classification, such as natural language inference and paraphrase detection. Some of these methods are unsupervised, and can be trained from large-scale unlabeled data~\cite{kiros2015skip,pgj2017unsup}. Other methods are trained from annotations of paraphrase, entailment, and contradiction among sentence pairs~\cite{wieting2016iclr,conneau2017supervised}, which seem related to at least some of the Penn Discourse Treebank relation types.

In this paper, we evaluate whether these sentence embedding techniques can improve performance on the classification of implicit discourse relations.
We use these techniques to embed each argument of the discourse relation, and then predict the relation type in a feedforward neural network. We compare these pre-trained sentence embedding models with a discriminatively-trained {Bi-LSTM} encoder, which is trained for the specific task of predicting discourse relations.

We find that supervised sentence embeddings like InferSent~\cite{conneau2017supervised}, which are trained on natural language inference tasks, tend to outperform unsupervised sentence embeddings like SkipThought~\cite{kiros2015skip}, which are trained on unlabeled data. The performance of these methods in comparison with the discriminative Bi-LSTM is mixed: the Bi-LSTM does well on the PDTB evaluation, in which both the training and test data consists of Wall Street Journal texts, but it suffers on the more recent CoNLL shared task data, in which the test set is drawn from newswire texts. This supports the hypothesis that pre-trained sentence embeddings can improve robustness to domain transfer. We also find that surface features are still important for this task, with the addition of simple word pair features improving the performance of all of the neural models. Replication code from these evaluations will be released upon publication of the paper.

%% file: data.tex
\section{Discourse Relation Prediction}

\paragraph{Dataset} 
There are now two well-studied datasets for discourse relation prediction in English: the Penn Discourse Treebank~\cite[PDTB;][]{rashmi2008penn} and the CoNLL 2016 shared task dataset~\cite{xue2015conll,xue2016conll}. The PDTB is a discourse level annotation of the Wall Street Journal (WSJ) articles. In this dataset, implicit relations are constrained by adjacency: only pairs of adjacent sentences within paragraphs are examined for the existence of implicit relations. The annotators selected only as much of the adjacent sentences as was minimally necessary for the interpretation of the inferred relation and these are called as arguments. These relation senses are arranged in a hierarchy, allowing for annotations as specific as \textit{Contingency.Cause.reason}. In our experiments, we use the second level of the implicit sense annotations as mentioned in \cite{rashmi2008penn}.

The training and development data for the CoNLL-2016 shared task was adapted from the same PDTB 2.0 corpus. However, they provide two test sets for the shared task: Section 23 of the PDTB, and a blind test set prepared especially for the shared task from English newswire texts.

\paragraph{Task} We focus on the problem to categorize second-level implicit discourse relations. Specifically, given the gold-standard arguments, 
the model needs to predict the correct sense of the implicit discourse relation for a given input. 
For example:  
\begin{itemize}
\itemsep0em
\item \textbf{Arg1:}\textit{ The brokerage firms learned a lesson the last time around.} 
\item \textbf{Arg2:}\textit{ This time, the firms were ready.}
\end{itemize}
The discourse relation here is \textit{Contingency.cause} indicating that the situation described in \textit{Arg1} influences the one in \textit{Arg2}.

%% file: methods.tex
\section{Incorporating Sentence Embeddings}

Without a semantic understanding of the sentences, we would not be able to guess the correct relation. The success of choosing the correct sense will require a representation that reflects the full meaning of a sentence. In this section, we describe all the models that we investigate for the sentence encoder. Specifically, we examine discriminatively-trained Bi-LSTM encoders trained for the specific task of predicting implicit discourse relations and unsupervised/pre-trained encoder models: Skipthought, Sent2Vec and Infersent and a combination of both the supervised and the pre-trained models. We obtain embeddings for both the sentence arguments, concatenate them and then train a feed forward neural network to classify the sense of the implicit discourse relation. 




\subsection{Supervised Embeddings} For a sequence of $T$ words $\{w_t\}^T_{t=1}$, a bidirectional LSTM consists of two LSTMs each reading the words in opposite directions. We explore three different ways to obtain sentence representations, concatenation of the forward and backward hidden states and mean/max-pooling of the hidden states. These models are trained end-to-end in a supervised manner to predict the discourse relations. 

\paragraph{Concatenation of Forward and Backward states}
We use the concatenation of the last vector of the Forward LSTM and the last vector of the Backward LSTM, which then forms a fixed-size vector representation of the sentence.
\begin{align*}
    & \overrightarrow{h_t} = \overrightarrow{LSTM}(w_1,...,w_T)_t \\
    & \overleftarrow{h_t} = \overleftarrow{LSTM}(w_1,...,w_T)_t \\
    & h_s = [\overrightarrow{h_T}, \overleftarrow{h_1}] 
\end{align*}

\paragraph{Pooling}
We also experiment with two additional ways of combining the varying number of hidden states $\{h_t\}$ to form a fixed-size vector, either by selecting the maximum value over each dimension of the hidden units (max pooling) \cite{Collobert:2011:NLP:1953048.2078186} or by considering the average of the representations (mean pooling).

\begin{align*}
    & \overrightarrow{h_t} = \overrightarrow{LSTM}(w_1,...,w_T)_t \\
    & \overleftarrow{h_t} = \overleftarrow{LSTM}(w_1,...,w_T)_t \\
	& h_t = [\overrightarrow{h_t}, \overleftarrow{h_t}] \\
	& h_s = \text{Max-Pool/Mean-Pool}(h_1,...,h_t) 
\end{align*}

\subsection{Pre-trained Embeddings}
To test the utility of pre-trained sentence embedding models for discourse relation classification, we apply three recent methods.

\paragraph{SkipThought}
Skipthought~\cite{kiros2015skip} is trained as part of an encoder-decoder model that tries to reconstruct the surrounding sentences of an encoded passage on the BookCorpus dataset~\cite{zhu2015aligning}.

\paragraph{Sent2Vec}
Sent2Vec~\cite{pgj2017unsup} is an unsupervised model trained on English wikipedia, which composes sentence embeddings from unigram and bigram embeddings. 

\paragraph{InferSent}
InferSent \cite{conneau2017supervised} provides a Bi-LSTM model with max-pooling that is trained on the natural language inference task~\cite{bowman2015large}. We use this encoder to obtain the sentence embeddings in our experiments.

\subsection{Combination}
We also try a combination of the sentence representations obtained from the pre-trained encoder and the end-to-end trained models. We concatenate the sentence embedding from the pre-trained encoder with the representation obtained from the Bi-LSTM model and then use this as the final representation of the sentence.

%% file: results.tex
\section{Experiments}
We evaluate our models on the test dataset of PDTB corpus and the CoNLL-2016 dataset. Similar to \cite{Pitler:2009:ASP:1690219.1690241}, we use sections 2-20 of the corpus as a training set, sections 0-1 as a development set for parameter tuning, and sections 21-22 for testing. As mentioned in \cite{Lin:2011:AET:2002472.2002598} around 2\% of the implicit relations in the PDTB are annotated with more than one type. During training, each argument pair that is annotated with two relation types is considered as two training instances, each with one relation type. During testing, if the classifier assigns either of the two types, it is considered to be correct. For CoNLL-16 dataset, We evaluate on both the publicly available and the blind test dataset.

\subsection{Experimental Settings}
We train all the models by minimizing the negative log likelihood of the correct relation for each pair of arguments in the training dataset. We have also tried training the model using a hinge loss and the results were similar or worse. Among the three different ways to obtain representations from the Bi-LSTM models trained end-to-end, simple concatenation of the forward and backward hidden states performs slightly better than both max and mean pooling of the hidden states. Hence, we excluded the other results.

For the pre-trained encoder model, the sentence vectors for the arguments are concatenated and then used as a base layer for a 4-layer feed forward neural network. In the end-to-end neural model, we use 300-dimensional Glove embeddings \cite{pennington2014glove} for the words as inputs to a 2-layered Bi-LSTM model with hidden state size set to 250 and the obtained sentence representation is used as the input to a 3-layer feed forward neural network. For a combined system, we use a 4-layer feed forward neural network for classification. The number of layers was chosen from \{2,3,4,5,7,10\}, with 3 or 4 fully connected layers giving best performance on the development data.

We use ReLU activation \cite{Nair:2010:RLU:3104322.3104425} for all the layers and use Xavier initialization for all the parameters \cite{pmlr-v9-glorot10a}. The model is trained using \textsc{Adam} optimizer~\cite{journals/corr/KingmaB14} with initial learning rate equal to 0.001 and dropout set to 0.35. 

\subsection{Results and Discussion}

\autoref{tab:tab1} summarizes the main empirical findings where we present the results of all our systems for the task of predicting second-level implicit discourse relations in PDTB test and both the CoNLL test datasets.

\begin{table*}
\small
\setlength{\tabcolsep}{4.5pt}
\begin{center}
\begin{tabular}{l|lll|ll|ll}
\toprule
& \multicolumn{3}{c|}{PDTB-test} & \multicolumn{2}{c|}{  CoNLL-blind test}& \multicolumn{2}{c}{CoNLL-test}\\
\toprule
  \textbf{Model} & \textbf{Acc.} & \textbf{+Bi-LSTM} & \textbf{+Bi-LSTM} & \textbf{Acc.} & \textbf{+Bi-LSTM} & \textbf{Acc.} & \textbf{+Bi-LSTM}\\
  &  &  &  \textbf{+WordPairs} &  &  & & \phantom{}\\
  \midrule
Bi-LSTM & \textbf{41.96\%} &  & 42.15\% &  31.55\% & & 36.03\% & \\[4pt]
Skipthought & 40.61\% & 43.02\% & \textbf{44.17}\% & 33.88\% & \textbf{36.47\%} & 34.23\% & 38.77\%\\
Sent2Vec & 37.63\% & 41.48\% & 43.02\% & 31.05\% & 34.11\% & 36.29\% & 36.94\% \\
Infersent & 40.51\% & \textbf{43.59\%} & 43.86\% & \textbf{34.64\%} & 36.0\% & \textbf{39.16\%} & \textbf{39.17\%} \\[4pt]
most common class & 25.11\% &  & & 17.64\%  &  & 25.64\% & \\
\cite{Lin:2009:RID:1699510.1699555} & 40.2\% &  &  &  &  &  & \phantom{}\\ 
\cite{Nie2017DisSentSR} & 42.9\%  & & & &  &  & \phantom{}\\ 
\cite{ji2015one} & \textbf{44.59\%}  & & & & & & \phantom{}\\
\cite{DBLP:journals/corr/QinZZHX17} & \textbf{46.23\%}  &  &  &  &  &  & \phantom{}\\
\bottomrule
\end{tabular}
\end{center}
\caption{\label{tab:tab1} Comparison of performance of all the models on the PDTB and CoNLL test datasets.}
\end{table*}

The most-common class is our baseline, where all instances are classified as \textit{Contingency.Cause}. While the performance of the pre-trained encoder models is similar to that of the end-to-end trained neural models, the Bi-LSTM model outperforms other models on the PDTB dataset and the Infersent encoder model is better on both the test datasets of CoNLL. 

A combined system of these two models outperforms individual models on both these datasets. All the models achieve a low performance on the blind test set, and the possible reason as mentioned in \cite{Wang2016TwoES} is that the CoNLL-blind test set has a different sense distribution compared with CoNLL-dev and CoNLL-test sets which are both from PDTB dataset, whereas the CoNLL-blind test set is annotated from English Wikinews. We also try adding the word-pair Brown cluster features for experiments on the PDTB dataset as mentioned in \cite{rutherford2014discovering} and observe improvements on all these systems thus showing that surface features are still crucial for this task. 
However, \cite{DBLP:journals/corr/QinZZHX17} still has the leading scores on PDTB. Note that their model is trained on both annotated relation sense and implicit connective information and hence, is complementary to our approach.

Upon inspection of the errors of these systems, interestingly we notice that all these models make similar kind of mistakes on the dataset. Given the huge overlap of these errors, it is clear that some instances are fundamentally more difficult and need additional document context for relation prediction.

%% file: relatedwork.tex
\section{Related Work}
Classical approaches to discourse relation classification are based on surface features such as word pairs~\cite{Marcu:2002:UAR:1073083.1073145,Pitler:2009:ASP:1690219.1690241} and syntax~\cite{lin2009recognizing}.
Feature selection was a particularly important characteristic of these approaches, as the number of word pair and syntactic features grows far faster than the amount of labeled data~\cite{Park:2012:IID:2392800.2392818,biran2013aggregated}
Neural network approaches were introduced first in the context of Rhetorical Structure Theory parsing~\cite{ji2014representation,li2014recursive}, and later in the Penn Discourse Treebank~\cite{ji2015one}. In general, these neural approaches are similar to the end-to-end Bi-LSTM model evaluated in this paper: they produce vector representations of each discourse relation argument, and then train a predictive model that takes these representations as input. \newcite{ji2015one} report that their recursive neural network architecture offers only modest improvements over the most competitive surface feature approaches~\cite{lin2009recognizing}, and similar results were obtained in the CoNLL shared tasks on discourse relation classification~\cite{rutherford2017systematic} . While earlier prior work has tested the effectiveness of pre-trained \emph{word} representations for discourse relation classification~\cite{rutherford2014discovering}, the impact of \emph{sentence} embeddings has not previously been evaluated.
Although some recent work \cite{Nie2017DisSentSR} was proposed to learn sentence representations by training a model to predict the discourse connective for explicit discourse relations and then evaluated on second-level implicit relation prediction task. We compare our results against theirs in \autoref{tab:tab1}.






%% file: conclusion.tex
\section{Conclusion}
In this paper, we focus on the task of predicting implicit discourse relations which requires obtaining rich semantic representations of the arguments.
Due to limited availability in training data and the semantic nature of the task, we explore various sentence encoding methods that can be trained on large datasets to obtain sentence representations. We show that the pre-trained sentence embedding methods are competitive with end-to-end trained neural models, and that a combination of these systems outperforms both the individual models. 

For future work, we plan on investigating why all of these systems do so poorly on this task. We are also particularly interested in domain adaptation based approaches to refine and obtain better semantic sentence representations for discourse relation prediction.